\documentclass{article}
\usepackage[latin9]{inputenc}
\usepackage{color}
\usepackage{float}
\usepackage{textcomp}
\usepackage{amsmath}
\usepackage{amssymb}
\usepackage{graphicx}
\usepackage[unicode=true,pdfusetitle,
 bookmarks=true,bookmarksnumbered=false,bookmarksopen=false,
 breaklinks=false,pdfborder={0 0 1},backref=false,colorlinks=false]
 {hyperref}

\makeatletter

\providecommand{\tabularnewline}{\\}
\floatstyle{ruled}
\newfloat{algorithm}{tbp}{loa}
\providecommand{\algorithmname}{Algorithm}
\floatname{algorithm}{\protect\algorithmname}

\usepackage{authblk}
\newtheorem{theorem}{Theorem}
\AtBeginDocument{\renewcommand{\ref}[1]{\mbox{\autoref{#1}}}}
\renewcommand*{\equationautorefname}[1]{}

\author{Venelin Mitov and Manfred Claassen}
\affil{Institute of Molecular Systems Biology, ETH Zurich}

\makeatother

\begin{document}

\title{A Fused Elastic Net Logistic Regression Model for Multi-Task Binary
Classification}
\maketitle
\begin{abstract}
Multi-task learning has shown to significantly enhance the performance
of multiple related learning tasks in a variety of situations. We
present the fused logistic regression, a sparse multi-task learning
approach for binary classification. Specifically, we introduce sparsity
inducing penalties over parameter differences of related logistic
regression models to encode similarity across related tasks. The resulting
joint learning task is cast into a form that lends itself to be efficiently
optimized with a recursive variant of the alternating direction method
of multipliers. We show results on synthetic data and describe the
regime of settings where our multi-task approach achieves significant
improvements over the single task learning approach and discuss the
implications on applying the fused logistic regression in different
real world settings.
\end{abstract}

\section{INTRODUCTION}

In this paper, we present fused logistic regression, a novel multi-task
learning algorithm for solving a set of binary classification tasks
that are ordered according to their mutual similarity. 

Multi-task learning refers to techniques that jointly address several
related learning tasks while leveraging their relatedness. Multi-task
learning has been applied in various settings. These include the recognition
of spam e-mails in different demographic groups \cite{Attenberg2009};
identification of host-pathogen protein interactions in different
infectious diseases \cite{Kshirsagar2013}; modeling of marketing
preferences of similar social groups \cite{Evgeniou2005a}. \textcolor{black}{A
widely adopted mechanism to take advantage of task relatedness is
to represent it in the form of a graph with weighted edges in which
every node is associated with an individual task or a group of related
tasks and to apply the adjacency matrix of this graph as a penalizing
multiplier for some metric of the difference between individual model
parameters \cite{Widmer2012a,Kim2010a}.} Multi-task linear regression
approaches exploit this (dis-) similarity graph by encouraging closely
related tasks to share a similar set of relevant input features using
structural-sparsity-inducing penalties \cite{Kim2010a,Chen2010,Yuan2006,Zhou2013a}. 

Sparsity-inducing penalties such as an L1-penalty on the model parameters
have been used to perform automatic variable selection \cite{Society2007,Liu2009,Clemmensen2008}.
\cite{Zou2005a} show that in linear regression, a combination of
an L1 and L2 penalty on the model parameters can lead to sparse model
fits while preserving or eliminating all parameters associated with
groups of strongly correlated predictors. This is particularly useful
in the case of small number of training samples, $n$, compared to
the dimensionality of the feature space $d$ ($n\ll d$), when the
L1 penalty alone has been shown to limit the number of selected features
to $n$ \cite{Zou2005a}. Different studies have shown that sparsity-inducing
priors can be applied as well in linear models for classification,
such as logistic regression and linear discriminant analysis.

Many real world learning tasks can be cast to the problem of learning
multiple related binary classification tasks. \textcolor{black}{For
example, in biology, correlation of time series of transcriptome profiles
with ordinal phenotypes or in marketing or in marketing, modeling
the buying preferences of different age groups of customers by associating
adjacent ranges of age with ordered classification tasks.}\textcolor{red}{{}
}Therefore, a need is present to develop models for multi-task binary
classification, which exploit task relatedness, while performing automatic
variable selection in a high-dimensional feature space. 

We propose the fused logistic regression to learn binary classifiers
in this situation. Specifically, we use (elastic net) logistic regression
models for the individual classification tasks. To leverage the similarity
across related tasks, we jointly fit all logistic regression models
while imposing L1 penalties on the parameter differences of related
tasks. The use of an elastic net penalty for each individual model
favors sparse estimates of the coefficient vectors while keeping groups
of correlated relevant predictor variables. 

In \ref{sec:LINEAR-LOGISTIC-REGRESSION} we establish the mathematical
notation for the rest of the paper and recapitulate the linear logistic
regression model for binary classification. In \ref{sec:THE-ADMM-METHOD}
we briefly describe the Alternating Direction Method of Multipliers
for convex optimization \cite{Boyd} of the learning objective induced
by the fused logistic regression model. In \ref{sec:THE-FUSED-ELASTIC},
we describe the fused elastic net logistic regression model for ordered
multi-task binary classification. In \ref{sec:EXPERIMENTS}, we report
experiments of the method conducted on synthetically generated data.

\section{LINEAR LOGISTIC REGRESSION FOR BINARY CLASSIFICATION\label{sec:LINEAR-LOGISTIC-REGRESSION}}

We start by a brief overview of single-task binary classification.
Given is training data which are realizations form $(X_{1},Y_{1}),...,(X_{n},Y_{n})\ \text{i.i.d.,}$
where the predictor or feature vector $X_{i}\subset\mathbb{R}^{d}$,
$i=1,...,n$, is a random vector and the vector of classes or labels
$Y=(Y_{1},...,Y_{n})\subset\{-1,1\}^{n}$ is a discrete random vector.
We denote the training data as the extended matrix $[X|\mathbf{y}]$,
where $X=[\mathbf{x}_{1},...,\mathbf{x}_{n}]^{T}\in\mathbb{R}^{n\times d}$
is called the design matrix, and the vector $\mathbf{y}=(y_{1},...,y_{n})\in\{-1,1\}^{n}$
is called the response vector. A classifier is a function, $\mathcal{C}:\mathbb{R}^{d}\rightarrow\{-1,1\}$,
assigning to a predictor vector $\mathbf{x}\in\mathbb{R}^{d}$ an
output label, which is a prediction for the corresponding true label
$y$. 

We consider linear logistic regression as our method of choice for
finding an estimator of the class probabilities $\pi_{y}(\mathbf{x}),\, y\in\{-1,1\}$.
While it performs comparably to competing methods, such as support
vector machines and linear discriminant analysis, logistic regression
has some notable advantages in that it provides a direct estimate
of $\pi_{y}(\mathbf{x})$ and tends to be more robust in the case
$d\gg n$ because it doesn't make assumptions on the distribution
of the predictors $X$. The logistic regression model is%
\footnote{To simplify the notation, we assume that we have added an intercept
element equal to $1$ as first component of the predictor vector $\mathbf{x}$.%
}
\begin{equation}
\hat{\pi}_{y}(\mathbf{x};\boldsymbol{\beta})=\sigma(y\mathbf{x}^{T}\boldsymbol{\beta})=\frac{1}{1+\exp(-y\mathbf{x}^{T}\boldsymbol{\beta})},\, y\in\{-1,1\}\label{eq:LLR}
\end{equation}
where $\boldsymbol{\beta}=(\beta_{0},\beta_{1},...,\beta_{d})^{T}$
is a vector of unknown parameters. The maximum likelihood estimate
for the parameters $\boldsymbol{\beta}$ is found by minimizing the
negative conditional log-likelihood function
\begin{eqnarray}
-\ell(\mathbf{\boldsymbol{\beta}}) & =\sum\log\left(\mathbf{1}+\exp(-\mathbf{y}\odot X\boldsymbol{\beta})\right),\label{eq:negLogLik}
\end{eqnarray}
where the symbol $'\sum'$ without subscript denotes sum over all
elements of the underlying vector or matrix, the symbol '$\odot$'
denotes the element-wise multiplication between vectors or matrices
with the same dimensions and the bold number $\mathbf{1}$ denotes
the $n$-dimensional real vector having all elements equal to $1$.
To improve the generalization performance of the model and to perform
automatic variable selection, we consider maximum a-posteriori estimates
of $\boldsymbol{\beta}$ by the use of a product of a Gaussian and
Laplacian densities, centered at $\mathbf{0}$ as a regularizing and
variable-selecting prior:
\[
p(\boldsymbol{\beta}_{\backslash0})={\cal N}\left(\mathbf{0},\frac{1}{\lambda_{2}}I\right)\times Lap\left(\mathbf{0},\frac{1}{\lambda_{1}}I\right),
\]
where $\lambda_{1}>0$ and $\lambda_{2}>0$, and by $\boldsymbol{\beta}_{\backslash0}$
we denote the vector $\boldsymbol{\beta}$ with omitted $0^{th}$
element. This results in the elastic net \cite{Zou2005a} estimate
of $\boldsymbol{\beta}$: 
\begin{equation}
\mathbf{\hat{\boldsymbol{\beta}}}_{EN}:=\arg\min_{\boldsymbol{\beta}\in\mathbb{R}^{1+d}}-\ell(\mathbf{\boldsymbol{\beta}})+\lambda_{1}||\boldsymbol{\beta}_{\backslash0}||_{1}+\frac{1}{2}\lambda_{2}||\boldsymbol{\beta}_{\backslash0}||_{2}^{2}\label{eq:MinNegLogLik}
\end{equation}

\section{THE ADMM METHOD FOR CONVEX OPTIMIZATION\label{sec:THE-ADMM-METHOD}}

A major challenge in statistical modeling is to define a model that,
on the one hand, is well adaptable to the phenomenon of study, and
on the other hand, can be fit to the training data in an efficient
way. In the case of model-fitting via likelihood or posterior maximization,
the fitting procedure reduces to an optimization problem. Linear models
like linear regression, logistic regression, linear discriminant analysis
and their L2- and L1- regularized variants are expressive and convenient
to fit to data since their learning objectives are convex. The Alternating
Direction Method of Multipliers (ADMM) is a convex optimization technique
that is particularly suited to optimize composite convex objectives,
as for instance the objective induced by the fused elastic net logistic
regression \ref{sec:THE-FUSED-ELASTIC}. 

The Alternating Direction Method of Multipliers (ADMM) \cite{Boyd}
is an optimization algorithm for solving constrained optimization
problems of the form:
\begin{equation}
\begin{aligned}\min\,\,\,\,\, f(\boldsymbol{\chi})+g(\boldsymbol{\zeta})\\
\text{subject to }\ensuremath{P\boldsymbol{\chi}+Q\boldsymbol{\zeta}=\mathbf{s}}
\end{aligned}
\label{eq:admmDecomposedProblem}
\end{equation}
with variables $\boldsymbol{\chi}\in\mathbb{R}^{n}$ and $\boldsymbol{\zeta}\in\mathbb{R}^{m}$,
where $P\in\mathbb{R}^{p\times n}$, $Q\in\mathbb{R}^{p\times m}$
and $\mathbf{s}\in\mathbb{R}^{p}$.%
\footnote{Here we use a slightly modified notation from the original paper \cite{Boyd}
with the Greek analogs of the the letters 'x' and 'z' in order to
avoid the conflict with the name 'x' for predictor variables.%
} 

The ADMM algorithm finds a saddle point of the augmented Lagrangian 

\begin{eqnarray*}
L_{\rho}(\boldsymbol{\chi},\boldsymbol{\zeta},\mathbf{w}) & := & f(\boldsymbol{\chi})+g(\boldsymbol{\zeta})+\boldsymbol{\omega}^{T}(P\boldsymbol{\chi}+Q\boldsymbol{\zeta}-\mathbf{s})\\
 &  & +\frac{1}{2}\rho||P\boldsymbol{\chi}+Q\boldsymbol{\zeta}-\mathbf{s}||_{2}^{2}.
\end{eqnarray*}
by iteratively solving smaller localized optimization tasks. With
the scaled dual variable $\boldsymbol{\xi}:=\boldsymbol{\omega}/\rho$,
\ref{alg:ADMM-(general-scaled} lists the general scaled form of ADMM
\cite{Boyd}.

\begin{algorithm}[h]
\caption{ADMM (general scaled form)\label{alg:ADMM-(general-scaled}}

Initialization: $\boldsymbol{\chi}^{0}=\boldsymbol{\zeta}^{0}=\boldsymbol{\xi}^{0}=\mathbf{0};\, k=0$\medskip{}

do \{

$\boldsymbol{\chi}$-update:\\
$\boldsymbol{\chi}^{k+1}:=\arg\min_{\boldsymbol{\chi}}\left(f(\boldsymbol{\chi})+\frac{1}{2}\rho||P\boldsymbol{\chi}+Q\boldsymbol{\zeta}^{k}-\mathbf{s}+\boldsymbol{\xi}^{k}||_{2}^{2}\right)$
\medskip{}

$\boldsymbol{\zeta}$-update:\\
$\boldsymbol{\zeta}^{k+1}:=\arg\min_{\boldsymbol{\zeta}}\left(g(\boldsymbol{\zeta})+\frac{1}{2}\rho||P\boldsymbol{\chi}^{k+1}+Q\boldsymbol{\zeta-\mathbf{s}}+\boldsymbol{\xi}^{k}||_{2}^{2}\right)$\medskip{}

$\boldsymbol{\xi}$-update: $\boldsymbol{\xi}^{k+1}:=\boldsymbol{\xi}^{k}+P\boldsymbol{\chi}^{k+1}+Q\boldsymbol{\zeta}^{k+1}-\mathbf{s}$\medskip{}

~$k=k+1$\medskip{}

\} while($k<MAXITER$ and not converged) 
\end{algorithm}

ADMM doesn't require strict convexity of its objective. This property
makes it a good candidate for solving problems with L1-norm terms
on the parameters, which have been shown to be not strictly convex
in the case $d\gg n$ \cite{Tibshirani2013}.

\section{THE FUSED ELASTIC NET LOGISTIC REGRESSION MODEL\label{sec:THE-FUSED-ELASTIC}}

We consider a set of $t$ ordered binary classification tasks, $1,...,t$,
on a set of $n$ $d$-dimensional labeled training observations $\{\mathbf{x}_{1},...,\mathbf{x}_{n}\}\subset\mathbb{R}^{1+d}$,
with $x_{i1}=1,\ i=1,..,n$. The order of the tasks reflects their
similarity. For instance, neighboring tasks should be more likely
to assign the same label to a test observation, compared to tasks
that are ordered far from each other. The training data for all tasks
is encoded in the matrix $[X|Y]$, where $X=[\mathbf{x}_{1},...,\mathbf{x}_{n}]^{T}\in\mathbb{R}^{n\times(1+d)}$
is the common design matrix shared by all tasks%
\footnote{Assume that the first column of the design matrix $X$ is the constant
vector $\mathbf{1}$.%
}, and the response vector for task $j$ is written as the column-vector
$\mathbf{y}_{j}=Y_{\cdot j}$ of the matrix $Y=[\mathbf{y}_{1},...,\mathbf{y}_{t}]\in\{-1,1\}{}^{n\times t}$,
$j=1,...,t$. We define a single-task logistic regression model for
task $j=1,...,t$ with training data $[X|\mathbf{y}_{j}]$ as:
\begin{equation}
\hat{\pi}_{y}^{(j)}(\mathbf{x};\boldsymbol{\beta}^{(j)})=\sigma(y\mathbf{x}^{T}\boldsymbol{\beta}^{(j)})=\frac{1}{1+\exp(-y\mathbf{x}^{T}\boldsymbol{\beta}^{(j)})},\label{eq:LLR_MTL}
\end{equation}
where $y\in\{-1,1\}$ and $\mathbf{\boldsymbol{\beta}}^{(j)}\in\mathbb{R}^{1+d}$
are the logistic regression parameters of task $j$. The negative
log-likelihood is defined in the same way as in (\ref{eq:negLogLik}):
\begin{equation}
-\ell^{(j)}(\mathbf{\boldsymbol{\beta}}^{(j)})=\sum\log\left[\mathbf{1}+\exp(-\mathbf{y}_{j}\odot X\mathbf{\boldsymbol{\beta}}^{(j)})\right],\label{eq:lcondshort1task}
\end{equation}
for $j=1,...,t$. As we saw in the introduction section, minimizing
an L1-L2-penalized version of the negative log-likelihood leads to
sparse solutions keeping non-zero parameters for the relevant sets
of correlated feature-vectors. This idea reduces to a single-task
fitting procedure, in which we find the L1-L2 penalized estimate of
the parameters by consecutively solving the optimization problems

\begin{equation}\begin{split}\boldsymbol{\beta}^{(j)*}:=\arg\min_{\boldsymbol{\beta}^{(j)}\in\mathbb{R}^{(1+d)}}\Big\{&-\ell^{(j)}(\mathbf{\boldsymbol{\beta}}^{(j)};[X|\mathbf{y}_{j}])+||\boldsymbol{\lambda}_{1}\odot\boldsymbol{\beta}^{(j)}||_{1}+\frac{1}{2}||\boldsymbol{\lambda}_{2}\odot\boldsymbol{\beta}^{(j)}||_{2}^{2}\Big\}\label{eq:L1L2LLR}\end{split}\end{equation}for
$j=1,..,t$. A small detail of this formulation is that we have presented
the regularizing parameters $\lambda_{1}>0$ and $\lambda_{2}>0$
as real vectors of the form $\boldsymbol{\lambda}_{1}=(0,\lambda_{1},...,\lambda_{1})\in\mathbb{R}^{1+d}$
and $\boldsymbol{\lambda}_{2}=(0,\lambda_{2},...,\lambda_{2})\in\mathbb{R}^{1+d}$,
in order to account for the usually unpenalized intercept $\beta_{0}^{(j)}$.

Now we wish to incorporate the prior knowledge about the similarity
between neighboring tasks into the model-fitting procedure. An important
observation, which directly follows from the continuity of the modeling
function in (\ref{eq:LLR_MTL}), is that two logistic regression models
operating on the same data would produce similar output if their parameters
were close. Therefore, similarity between neighboring logistic regression
models for neighboring tasks can be encoded by penalizing the difference
between their parameters. Let $B:=[\mathbf{\boldsymbol{\beta}}^{(1)},...,\mathbf{\boldsymbol{\beta}}^{(t)}]\in\mathbb{R}^{(1+d)\times t}$
be the coefficient matrix for all tasks and let $R\in\mathbb{R}^{t\times t}$
be a matrix defined in the following way:
\[
R_{ij}:=\begin{cases}
1 & \text{if }j=i-1\\
0 & \text{otherwise}\\
Circ & \text{ if }(i,j)=(1,t)
\end{cases},\,\, i,j=1,...,t.
\]
The variable $Circ$ is equal to $0$ if no relatedness between task
$t$ and task $1$ should be modeled and to $1$, otherwise. We call
$R$ the column-rotating matrix for $B$ because the columns of the
$(1+d)\times t$-matrix $BR$ are the same as the columns of $B$,
but rotated by one column to the left. 

Let $\nu\geq0$ be a penalizing parameter. Denote by $[\mathbf{\cdot}]$
the $(1+d)\times t$-matrix with all columns equal to a vector $\mathbf{\cdot}$,
by $[\nu]$ the $(1+d)\times t$-matrix, each element of which is
equal to $\nu$, and by $I$ the $t\times t$-dimensional identity
matrix. We define the multi-task fused L1-L2-penalized negative log-likelihood
as the function:
\begin{eqnarray}
\ell^{MT}(B):=-\sum_{j=1}^{t}\ell^{(j)}(\mathbf{\boldsymbol{\beta}}^{(j)})+\sum_{j=1}^{t}||\boldsymbol{\lambda}_{1}\odot\boldsymbol{\beta}^{(j)}||_{1}\nonumber \\
+\sum_{j=1}^{t}\frac{1}{2}||\boldsymbol{\lambda}_{2}\odot\boldsymbol{\beta}^{(j)}||_{F}^{2}+||[\nu]\odot B(I-R)||_{1}\nonumber \\
=\sum\log\Bigl([1]+\exp(-Y\odot XB)\Bigr)+||[\boldsymbol{\lambda}_{1}]\odot B||_{1}\nonumber \\
+\frac{1}{2}||[\boldsymbol{\lambda}_{2}]\odot B||_{F}^{2}+||[\nu]\odot B(I-R)||_{1}\label{eq:fL1L2LLR_mat_fused}
\end{eqnarray}
The first equality shows that if the penalizing parameter $\nu$ is
set to $0$, the optimization can be split across the columns of $B$,
and is equivalent to the single-task optimization with elastic net
penalty (\ref{eq:MinNegLogLik}). The fusing L1 penalty (\ref{eq:fL1L2LLR_mat_fused})
represents a scaled sum of absolute differences between each pair
of consecutive columns of $B$ and cannot be decomposed column-wise.
The MAP fit of the parameters $B$ to the training data $[X|Y]$ is
found by solving the optimization problem
\begin{equation}
B^{*}=\arg\min_{B\in\mathbb{R}^{(1+d)\times t}}\ell^{MT}(B).\label{eq:minLogLikMT}
\end{equation}
As a sum of convex functions, the function $\ell^{MT}$ is also convex.
Through the rest of this section, we show one way to solve this problem
(\ref{eq:minLogLikMT}) reformulating it in an ADMM compliant form.

To begin, we convert problem (\ref{eq:minLogLikMT}) to the canonical
ADMM-form (\ref{eq:admmDecomposedProblem}) by introducing the variable
matrices $\chi\in\mathbb{R}^{(1+d)\times t}$ and $\zeta\in\mathbb{R}^{(1+d)\times t}$,
and separating the differentiable from the non-differentiable terms
as follows:
\[
f(\chi):=\sum\log\Bigl([1]+\exp(-Y\odot X\chi)\Bigr)+\frac{1}{2}||[\boldsymbol{\lambda}_{2}]\odot\chi||_{F}^{2},
\]
\[
g(\zeta):=||[\boldsymbol{\lambda}_{1}]\odot\zeta||_{1}+||[\nu]\odot\zeta(I-R)||_{1}.
\]

With this split, the canonical ADMM form for problem (\ref{eq:minLogLikMT})
is:

\begin{eqnarray}
\min &  & f(\chi)+g(\zeta)\label{eq:ADMMCanonicalFormForFLLR}\\
\text{subject to} &  & \chi-\zeta=[0].\nonumber 
\end{eqnarray}

The scaled form of the ADMM algorithm for problem (\ref{eq:ADMMCanonicalFormForFLLR})
is given in \ref{alg:ADMM-for lMT}:
\begin{algorithm}[H]
\caption{ADMM for $\ell^{MT}$\label{alg:ADMM-for lMT}}

\medskip{}
Initialization: $\chi^{0}=\zeta^{0}=\xi^{0}=[0]_{(1+d)\times t}$;$k:=0$\medskip{}

do \{

$\chi$-update:\\
$\chi^{k+1}:=\arg\min_{\chi}\left(f(\chi)+\frac{1}{2}\rho||\chi-\zeta^{k}+\xi^{k}||_{2}^{2}\right)$
\medskip{}

$\zeta$-update:\\
$\zeta^{k+1}:=\arg\min_{\zeta}\left(g(\zeta)+\frac{1}{2}\rho||\chi^{k+1}-\zeta+\xi^{k}||_{2}^{2}\right)$\medskip{}

$\xi$-update:\\
$\xi^{k+1}:=\xi^{k}+\chi^{k+1}-\zeta^{k+1}$\\

$k:=k+1$

\} while($k<MAXITER$ and not converged )
\end{algorithm}

The convergence criterion is straightforward to implement, following
the instructions in \cite[p. 16-17]{Boyd}. 

In the next two subsections, we describe the $\chi$-update and the
$\zeta$-update.

\subsection{Newton-Raphson Gradient Descent Procedure for the $\chi$-update\label{sub:Newton-Raphson-gradient-descent}}

For the $\chi$-update, we notice that there is no coupling between
the columns of the variable matrix $\chi$. Therefore, it is computationally
more convenient to obtain $\chi^{k+1}$ by solving separately for
$j=1,...,t:$
\begin{equation}
\chi_{\cdot j}^{k+1}:=\arg\min_{\chi_{\cdot j}}\,\tilde{f}^{\, k(j)}(\chi_{\cdot j}),\label{eq:x-update}
\end{equation}
where we denoted
\begin{eqnarray*}
\tilde{f}^{\, k(j)}(\chi_{\cdot j}) & := & \sum\log\Bigl(\mathbf{1}+\exp(-Y_{\cdot j}\odot X\chi_{\cdot j})\Bigr)\\
 &  & +\frac{1}{2}||\boldsymbol{\lambda}_{2}\odot\chi_{\cdot j}||_{2}^{2}+\frac{1}{2}\rho||\chi_{\cdot j}-(\zeta_{\cdot j}^{k}-\xi_{\cdot j}^{k})||_{2}^{2}.
\end{eqnarray*}

The function $\tilde{f}^{\, k(j)}(\chi_{\cdot j})$ is twice differentiable
and convex and, therefore, can be optimized efficiently using the
Newton-Raphson's method.

The Newton-Raphson method involves the evaluation of the inverse Hessian
of the objective. Due to its possibly large dimensionality, $(d\times d)$,
this step can become prohibitively expensive. Using the conventional
method ``solve'' in R, on a computer with 64-bit 3.1 GHz Intel\texttrademark{}
(Core\texttrademark{}) i7-processor, the inversion of the Hessian
matrix for $d=5000$ takes $\sim160s$, compared to $\sim1s$ for
$d=1000$, and $\sim0.15s$ for $d=500$. If we ignore the costs for
calculating the gradient and the hessian, with $d=5000$, a full ten
iteration Newton-Raphson's execution would take approximately $26$
minutes. This computational cost is prohibitive, considering that
this procedure will be repeated for each ADMM iteration. 

It turns out, that we can solve problem (\ref{eq:x-update}) by only
considering tractable inversions of $n-$dimensional matrices \cite{VanHouwelingen2006,Goeman2010}.
To simplify the notation, we denote $\Omega_{\cdot j}^{k}:=(\zeta_{\cdot j}^{k}-\xi_{\cdot j}^{k})$.
The gradient of $\tilde{f}^{\, k(j)}$ is
\begin{equation}
\nabla\tilde{f}^{\, k(j)}(\chi_{\cdot j})=X^{T}\delta(\chi_{\cdot j})+\eta(\chi_{\cdot j}),\label{eq:gradf}
\end{equation}
where we denoted $\delta(\chi_{\cdot j}):=\left[-Y_{\cdot j}\odot\exp(-Y_{\cdot j}\odot X\chi_{\cdot j})\right]\div\left[\mathbf{1+}\exp(-Y_{\cdot j}\odot X\chi_{\cdot j})\right],$
$\eta(\chi_{\cdot j}):=(\boldsymbol{\lambda}_{2}+\mathbf{\boldsymbol{\rho}})\odot\chi_{\cdot j}-\rho\Omega_{\cdot j}^{k}$
and the symbol $'\div'$ denotes element-wise division between its
vector or matrix operands and $I$ denotes the identity matrix.

We know that at the global minimum $\chi_{\cdot j}^{*}$ of $\tilde{f}^{\, k(j)}$
the gradient (\ref{eq:gradf}) should vanish. Setting the gradient
to the vector $\mathbf{0}$ reveals that there exists an $n$-dimensional
real vector $\boldsymbol{\gamma}_{j}^{*}:=-\delta(\chi_{\cdot j}^{*})$,
such that
\begin{eqnarray}
X^{T}\boldsymbol{\gamma}_{j}^{*} & = & \eta(\chi_{\cdot j}^{*})=(\boldsymbol{\lambda}_{2}+\mathbf{\boldsymbol{\rho}})\odot\chi_{\cdot j}^{*}-\rho\Omega_{\cdot j}^{k}.\label{eq:gammastar}
\end{eqnarray}
The two equations below follow directly from (\ref{eq:gammastar}):
\begin{equation}
\chi_{\cdot j}^{*}=(X^{T}\boldsymbol{\gamma}^{*}+\rho\Omega_{\cdot j}^{k})\div(\boldsymbol{\lambda}_{2}+\mathbf{\boldsymbol{\rho}})\label{eq:XstarOfGammaStar}
\end{equation}

\begin{equation}
\boldsymbol{\gamma}_{j}^{*}=(XX^{T})^{-1}X\left((\boldsymbol{\lambda}_{2}+\mathbf{\boldsymbol{\rho}})\odot\chi_{\cdot j}^{*}-\rho\Omega_{\cdot j}^{k}\right)\label{eq:gammaStarOfXstar}
\end{equation}
Equation (\ref{eq:XstarOfGammaStar}) shows that $\chi_{\cdot j}^{*}$
lies in an $n$-dimensional space. Let $h:\,\mathbb{R}^{n}\rightarrow\mathbb{R}^{d}$
and $h^{-1}:\,\mathbb{R}^{d}\rightarrow\mathbb{R}^{n}$ be the following
two (mutually inverse) functions: 
\begin{eqnarray}
h(\boldsymbol{\gamma}): & = & (X^{T}\boldsymbol{\gamma}+\rho\Omega_{\cdot j}^{k})\div(\boldsymbol{\lambda}_{2}+\mathbf{\boldsymbol{\rho}})\label{eq:hOfGamma}
\end{eqnarray}
\begin{eqnarray}
h^{-1}(\boldsymbol{\chi}): & = & (XX^{T})^{-1}X\left((\boldsymbol{\lambda}_{2}+\mathbf{\boldsymbol{\rho}})\odot\boldsymbol{\chi}-\rho\Omega_{\cdot j}^{k}\right)\label{eq:h_1OfX}
\end{eqnarray}

The following theorem will form the basis of defining an optimization
problem over an $n-$dimensional variable whose optimum can be used
to unambiguously reconstruct the $d-$dimensional solution of the
initial problem. 

\begin{theorem}
\label{thm:monotonictransform}
Let the function $\phi^{k(j)}:\,\mathbb{R}^{n}\rightarrow\mathbb{R}$ be defined as: \[ \phi^{k(j)}(\boldsymbol{\gamma}):=\tilde{f}^{\, k(j)}(h(\boldsymbol{\gamma})). \]
$\boldsymbol{\chi}^{*}$ is the global minimum of $\tilde{f}^{\, k(j)}$ if and only if  $\boldsymbol{\gamma}^{*}:=h^{-1}(\boldsymbol{\chi}^{*})$ is the global minimum of $\phi^{k(j)}$. 
\end{theorem}

It follows from \ref{thm:monotonictransform} that the minimization
problem (\ref{eq:x-update}) can be solved by minimizing the $n$
-dimensional function $\phi^{k(j)}$, and setting
\begin{equation}
\chi_{\cdot j}^{k+1}:=h\left(\arg\min_{\boldsymbol{\gamma}\in\mathbb{R}^{n}}\phi^{k(j)}(\boldsymbol{\gamma})\right).\label{eq:XfromPhi(Gamma)}
\end{equation}

Minimizing the function $\phi^{k(j)}$ is done again by the Newton-Raphson's
method without suffering from the costly inversion of a $d\times d$
-matrix. Analytical expressions for the gradient and hessian of $\phi^{k(j)}$
are provided in the appendix.

\subsection{Second-level ADMM for the $\zeta$-update\label{sub:Second-level-ADMM}}

In the sequel of this section we will rely on a fact, known from subdifferential
calculus \cite{ConvexAnalysis}. For $\kappa\in\mathbb{R}$, $\kappa\geq0$
and any real number $a$, the \textit{soft thresholding operator,}
$S_{\kappa}$, is defined as:
\[
S_{\kappa}(a):=\begin{cases}
a-\kappa & a>\kappa\\
0 & |a|\leq\kappa\\
a+\kappa & a<-\kappa.
\end{cases}
\]

For the $\chi$-update we use the following result \cite{Boyd}:
\begin{description}
\item [{Soft~thresholding:}] \label{thm:softthresholding}
Let $\lambda,\rho>0$, $x$ is a real variable and $v$ is some real constant. The optimiaztion problem \[ x^{*}:=\arg\min_{x}\left(\lambda|x|+(\rho/2)(x-v)^{2}\right) \] has the closed-form solution \[ x=S_{\lambda/\rho}(v). \]

\end{description}
The $\zeta$-update is: \begin{equation}\begin{split} \zeta^{k+1}:=&\arg\min_{\zeta}\Big\{||[\boldsymbol{\lambda}_{1}]\odot\zeta||_{1}\\&+||[\nu]\odot\zeta(I-R)||_{1}+\frac{1}{2}\rho||\zeta-\Omega||_{2}^{2}\Big\},\label{eq:z-update} \end{split}\end{equation}where
$\boldsymbol{\lambda}_{1}\in\mathbb{R}^{1+d},$ $\boldsymbol{\nu}\in\mathbb{R}^{1+d}$,
$\zeta,\,\chi^{k+1},\,\xi^{k}\in\mathbb{R}^{(1+d)\times t}$, $(I-R)\in\mathbb{R}^{t\times t}$,
$[\cdot]\in\mathbb{R}^{(1+d)\times t}$ denotes the matrix with $t$
columns, equal to the $(1+d)$-dimensional vector $\cdot$, and $\Omega:=\chi^{k+1}+\xi^{k}$.
Due to the two L1-norms, the objective function, unlike the case in
the $\chi$-update, is not column-wise decomposable. Again, we use
ADMM, to solve problem (\ref{eq:z-update}). Because the objective
function remains invariant with respect to transposition, we can write
problem (\ref{eq:z-update}) as:

\begin{equation}\begin{split} (\zeta^{k+1})^{T}:=&\arg\min_{\zeta^{T}}\Big\{||[\boldsymbol{\lambda}_{1}]^{T}\odot\zeta^{T}||_{1}\\&+||[\nu]^{T}\odot(I-R)^{T}\zeta^{T}||_{1}+\frac{1}{2}\rho||\zeta^{T}-\Omega^{T}||_{2}^{2}\Big\}\label{eq:z-update-transpone} \end{split}\end{equation}

Defining the two variables $\overset{\sim}{\chi}:=\zeta^{T}$ and
$\overset{\sim}{\zeta}:=(I-R)^{T}\zeta^{T}$, we present (\ref{eq:z-update-transpone})
in the canonical ADMM form \ref{eq:admmDecomposedProblem} as:

\begin{equation}
\begin{aligned}\min\,\,\,\,\,\overset{\sim}{f}(\overset{\sim}{\chi})+\overset{\sim}{g}(\overset{\sim}{\zeta})\\
\text{subject to }(I-R)^{T}\overset{\sim}{\chi}-\overset{\sim}{\zeta}\ensuremath{=[0]_{t\times(1+d)}},
\end{aligned}
\label{eq:admmDecomposedProblem-1}
\end{equation}
where $\overset{\sim}{f}(\overset{\sim}{\chi}):=||[\boldsymbol{\lambda}_{1}]^{T}\odot\overset{\sim}{\chi}||_{1}+\frac{1}{2}\rho||\overset{\sim}{\chi}-\Omega^{T}||_{2}^{2}$
, and $\overset{\sim}{g}(\overset{\sim}{\zeta}):=||[\nu]^{T}\odot\overset{\sim}{\zeta}||_{1}$.
The scaled-form ADMM for problem (\ref{eq:admmDecomposedProblem-1})
is given in \ref{alg:ADMM-(general-scaled-1} 
\begin{algorithm}[h]
\caption{ADMM for the $\zeta$-update\label{alg:ADMM-(general-scaled-1}}

Initialization: $\overset{\sim}{\chi}^{0}=\overset{\sim}{\zeta}^{0}=\overset{\sim}{\xi}^{0}=[0]_{t\times(1+d)};\, k=0$

do \{

$\overset{\sim}{\chi}$-update:\\
$\overset{\sim}{\chi}^{k+1}:=\arg\underset{\overset{\sim}{\chi}}{\min}\left(\overset{\sim}{f}(\overset{\sim}{\chi})+\frac{1}{2}\overset{\sim}{\rho}||(I-R)^{T}\overset{\sim}{\chi}-\overset{\sim}{\zeta}^{k}+\overset{\sim}{\xi}^{k}||_{2}^{2}\right)$
\medskip{}

$\overset{\sim}{\zeta}$-update:\\
$\overset{\sim}{\zeta}^{k+1}:=\arg\underset{\overset{\sim}{\zeta}}{\min}\left(\overset{\sim}{g}(\overset{\sim}{\zeta})+\frac{1}{2}\overset{\sim}{\rho}||(I-R)^{T}\overset{\sim}{\chi}^{k+1}-\overset{\sim}{\zeta}+\overset{\sim}{\xi}^{k}||_{2}^{2}\right)$\medskip{}

$\overset{\sim}{\xi}$-update:\\
$\overset{\sim}{\xi}^{k+1}:=\overset{\sim}{\xi}^{k}+(I-R)^{T}\overset{\sim}{\chi}^{k+1}-\overset{\sim}{\zeta}^{k+1}$\medskip{}

~$k=k+1$

\} while($k<MAXITER$ and not converged) 
\end{algorithm}

\subsubsection*{Iterative Soft Thresholding for the $\protect\overset{\sim}{\chi}$-update}

The $\overset{\sim}{\chi}$-update is:

\begin{equation}\begin{split} \overset{\sim}{\chi}^{k+1}:=&\arg\min_{\overset{\sim}{\chi}}\Big\{||[\boldsymbol{\lambda}_{1}]^{T}\odot\overset{\sim}{\chi}||_{1}+\frac{1}{2}\rho||\overset{\sim}{\chi}-\Omega^{T}||_{2}^{2}\\&+\frac{1}{2}\overset{\sim}{\rho}||(I-R)^{T}\overset{\sim}{\chi}-\overset{\sim}{\zeta}^{k}+\overset{\sim}{\xi}^{k}||_{2}^{2}\Big\}\label{eq:chicirc-update} \end{split}\end{equation}

We notice that the problem (\ref{eq:chicirc-update}) is column-wise
decomposable, meaning that we can split it into subproblems of the
form

\begin{equation}\begin{split} \overset{\sim}{\chi}_{\cdot l}^{k+1}:=&\arg\min_{\overset{\sim}{\chi}_{\cdot l}}\Big\{||[\boldsymbol{\lambda}_{1}]_{\cdot l}^{T}\odot\overset{\sim}{\chi_{\cdot l}}||_{1}+\frac{1}{2}\rho||\overset{\sim}{\chi}_{\cdot l}-\Omega_{\cdot l}^{T}||_{2}^{2}\\&+\frac{1}{2}\overset{\sim}{\rho}||(I-R)^{T}\overset{\sim}{\chi}_{\cdot l}-\overset{\sim}{\zeta}_{\cdot l}^{k}+\overset{\sim}{\xi}_{\cdot l}^{k}||_{2}^{2}\Big\}\label{eq:chicirc-column-update} \end{split}\end{equation}

The term $(I-R)^{T}\overset{\sim}{\chi}$ represents the $t\times(1+d)$-dimensional
matrix with $j^{\text{th}}$ row representing the row-vector difference%
\footnote{In the case $j=t$, $j+1$ should be thought of as $1$.%
} $\left(\overset{\sim}{\chi}_{j\,\cdot}-\overset{\sim}{\chi}_{(j+1)\,\cdot}\right)$.
Because of this coupling between consecutive rows of $\overset{\sim}{\chi}$,
problem \ref{eq:chicirc-update} cannot be row-decomposed. Therefore,
we use a coordinate descent approach for solving problem (\ref{eq:chicirc-column-update})
for $l=1,...,1+d$. 

Let $\overset{\sim}{\chi}_{\cdot l}$ be the current estimate of $\overset{\sim}{\chi}_{\cdot l}^{k+1}$
from (\ref{eq:chicirc-column-update}), and let $j\in\{1,...,t\}$.
Denote $\overset{\sim}{\Omega}_{(j-1)l}:=\bigl(\overset{\sim}{\chi}_{(j-1)l}-\overset{\sim}{\zeta}_{(j-1)l}^{k}+\overset{\sim}{\xi}_{(j-1)l}^{k}\bigr)$
and $\overset{\sim}{\Omega}_{jl}:=\bigl(\overset{\sim}{\chi}_{(j+1)l}+\overset{\sim}{\zeta}_{jl}^{k}-\overset{\sim}{\xi}_{jl}^{k}\bigr)$.
A coordinate descent step for the $j^{\text{th}}$ element of $\overset{\sim}{\chi}_{\cdot l}$
consists in solving
\begin{eqnarray*}
\overset{\sim}{\chi}_{jl}^{+} & := & \arg\min_{\overset{\sim}{\chi}_{jl}}\big\{\,||\lambda_{1l}\overset{\sim}{\chi_{jl}}||_{1}+\frac{1}{2}\rho||\overset{\sim}{\chi}_{jl}-\Omega_{lj}||_{2}^{2}\\
 &  & +\frac{1}{2}\overset{\sim}{\rho}||-\overset{\sim}{\chi}_{jl}+\overset{\sim}{\Omega}_{(j-1)l}||_{2}^{2}+\frac{1}{2}\overset{\sim}{\rho}||\overset{\sim}{\chi}_{jl}-\overset{\sim}{\Omega}_{jl}||_{2}^{2}\,\big\}\\
 & = & \arg\min_{\overset{\sim}{\chi}_{jl}}\bigg\{\,\lambda_{1l}|\overset{\sim}{\chi_{jl}}|+\frac{\rho+2\overset{\sim}{\rho}}{2}\,\biggl(\overset{\sim}{\chi}_{jl}-\frac{\rho\Omega_{lj}+\overset{\sim}{\rho}\overset{\sim}{\Omega}_{(j-1)l}+\overset{\sim}{\rho}\overset{\sim}{\Omega}_{jl}}{\rho+2\overset{\sim}{\rho}}\biggr)^{2}\bigg\}.
\end{eqnarray*}
By denoting $\kappa_{jl}:=\frac{\lambda_{1l}}{\rho+2\overset{\sim}{\rho}}$
and $a_{jl}:=\frac{\rho\Omega_{lj}+\overset{\sim}{\rho}\overset{\sim}{\Omega}_{(j-1)l}+\overset{\sim}{\rho}\overset{\sim}{\Omega}_{jl}}{\rho+2\overset{\sim}{\rho}}$
and using soft thresholding, we find: 
\[
\overset{\sim}{\chi}_{jl}^{+}=S_{\kappa_{jl}}(a_{jl}).
\]
To find $\overset{\sim}{\chi}_{\cdot l}^{k+1}$, we repeat the same
step, letting the index $j$ to iterate cyclically over the $\{1,...,t\}$
until satisfying a convergence criterion for the difference in the
objective function between two complete cycles.

\subsubsection*{Soft Thresholding for the $\protect\overset{\sim}{\zeta}$-update}

The $\overset{\sim}{\zeta}$-update is:
\begin{eqnarray}
\overset{\sim}{\zeta}^{k+1}: & = & \arg\min_{\overset{\sim}{\zeta}}\,\bigg\{\,||[\nu]^{T}\odot\overset{\sim}{\zeta}||_{1}\nonumber \\
 &  & +\frac{1}{2}\overset{\sim}{\rho}||(I-R)^{T}\overset{\sim}{\chi}^{k+1}-\overset{\sim}{\zeta}+\overset{\sim}{\xi}^{k}||_{2}^{2}\,\bigg\}.\label{eq:zetacirc-update}
\end{eqnarray}

This problem is column- and row-decomposable and can easily be solved
for each element of $\overset{\sim}{\zeta}^{k+1}$ by soft thresholding:
\begin{equation}
\overset{\sim}{\zeta}_{jl}^{k+1}=S_{\nu/\overset{\sim}{\rho}}\left(\overset{\sim}{\chi}_{jl}^{k+1}-\overset{\sim}{\chi}_{(j+1)l}^{k+1}+\overset{\sim}{\xi}_{jl}^{k}\right),
\end{equation}
where $j=\{1,...,t\},\, l=\{1,...,1+d\}.$

\section{EXPERIMENTS\label{sec:EXPERIMENTS}}

We designed comparative benchmarks with synthetic data-sets in order
to evaluate the following models:
\begin{description}
\item [{(i)~Fused~Elastic~Net~Logistic~Regression:}] the most general
model in which all regularizing parameters are allowed to be non-zero:$\lambda_{1}\geq0$,
$\lambda_{2}\geq0$, $\nu\geq0$;
\item [{(ii)~Fused~L1~Logistic~Regression:}] $\lambda_{1}\geq0$, $\lambda_{2}=0$,
$\nu\geq0$;
\item [{(iii)~Elastic~Net~Logistic~Regression:}] $\lambda_{1}\geq0$,
$\lambda_{2}\geq0$, $\nu=0$;
\item [{(iv)~Unpenalized~Logistic~Regression:}] $\lambda_{1}=0$, $\lambda_{2}=0$,
$\nu=0$;
\item [{(v)~Discrete~AdaBoost:}] Default implementation from the R-package
``ada'' \cite{Culp2006} with exponential loss functions and 200
iterations ;
\end{description}
Our benchmarks on synthetic data simulate different conditions with
respect to the correlation between feature vectors, the sparsity of
the model parameters and the degree of similarity between neighboring
tasks. Specifically, we simulated an ``Independent Features'' and
a ``Correlated Features'' scenario and for each of these we generated
quartets of 100-dimensional logistic regression coefficient vectors
with associated predictor data-sets simulating four sparsity and similarity
conditions as shown in \ref{tab:Simulated-Sparsity-and}. We measure
sparsity in terms of number of non-zero model parameters and similarity
in terms of number of matching non-zero parameters between two tasks. 

\begin{table}[h]
\caption{Simulated Feature Correlation, Sparsity and Similarity Conditions\label{tab:Simulated-Sparsity-and}}
\medskip{}

\begin{centering}
\begin{tabular}{lll}
\textbf{\footnotesize{1. Independent~~}} & \textbf{\footnotesize{Low similarity}} & \textbf{\footnotesize{High similarity}}\tabularnewline
\textbf{\footnotesize{Non-sparse}} & \textcolor{black}{\small{a) {[}60, 12{]}}} & \textcolor{black}{\small{b. {[}60, 48{]}}}\tabularnewline
\textbf{\footnotesize{Sparse}} & c) {[}10, 2{]} & \textcolor{black}{\small{d. {[}10, 8{]}}}\tabularnewline
\end{tabular}\medskip{}

\par\end{centering}

\begin{centering}
\begin{tabular}{lll}
\textbf{\footnotesize{2. Correlated~~}} & \textbf{\footnotesize{Low similarity}} & \textbf{\footnotesize{High similarity}}\tabularnewline
\textbf{\footnotesize{Non-sparse}} & \textcolor{black}{\small{e) {[}60, 12{]}}} & \textcolor{black}{\small{f) {[}60, 48{]}}}\tabularnewline
\textbf{\footnotesize{Sparse}} & g) {[}10, 2{]} & \textcolor{black}{\small{h) {[}10, 8{]}}}\tabularnewline
\end{tabular}\medskip{}

\par\end{centering}

{\small{In brackets are denoted the number of non-zero parameters
per task and the number of matching non-zero parameters between neighboring
tasks.}}
\end{table}
For each case (a-h) we generated 20 independent quartet/data-set instances
resulting in a total of 80 instances. Each quartet represents the
true logistic regression model parameters for four linearly ordered
binary classification tasks. The non-zero parameters are sampled from
the set $\{-4,\,-2,\,2,\,4\}$, while ensuring the case specific degrees
of sparsity and similarity. We use the term ``relevant feature''
to distinguish a feature for which at least one coefficient in at
least one task is non-zero. In scenario 1, ``Independent Features'',
the feature vectors are drawn from a standard normal distribution:
$\mathbf{x}\sim{\cal N}(\mathbf{0},\, I_{100})$ and are classified
randomly according to the logistic model (\ref{eq:LLR}). In scenario
2, ``Correlated Features'', the parameters and the feature vectors
are first drawn like in scenario 1. Then, the second $d/2$ parameters
are assigned the same values as their corresponding parameters in
the first half, while the relevant feature vectors in the second half
are re-sampled from a normal distribution $X_{\cdot(d/2+j)}\sim{\cal N}(X_{.(j)},\,0.4I_{d})$
so that they represent noisy copies of their corresponding relevant
feature vectors in the first half. This procedure guarantees positive
correlation in the order of 0.9 for couples of relevant feature vectors
corresponding to equal parameters. For each instance of each case
(a-h) we trained the five models (i-v) on data-sets ranging from 25
to 400 training observations. To tune the penalizing parameters $\lambda_{1}$,
$\lambda_{2}$ and $\nu$, we fitted the model-instances using parameter
combinations from the Cartesian product of $\Lambda_{1}=\{0,0.1,0.2,0.4,0.6,0.8,1,2,4,6,8\}$,
$\Lambda_{2}=\{0,0.05,0.1,0.2,0.4,1,2\}$, {\Large{$\nu$}}$=\{0,0.05,0.1,0.2,0.4,0.6,0.8,1,2,4,6,8\}$
and we used validation data-sets of 1400 observations to estimate
and compare the expected prediction errors. Finally, we evaluated
the expected $L_{01}$ error of all five models on separate data-sets
of 1400 observations that haven't been used either for the training
nor for the parameter tuning.

\subsection{Comparison of Model Predictive Performance}

\begin{figure}
\includegraphics[scale=0.75]{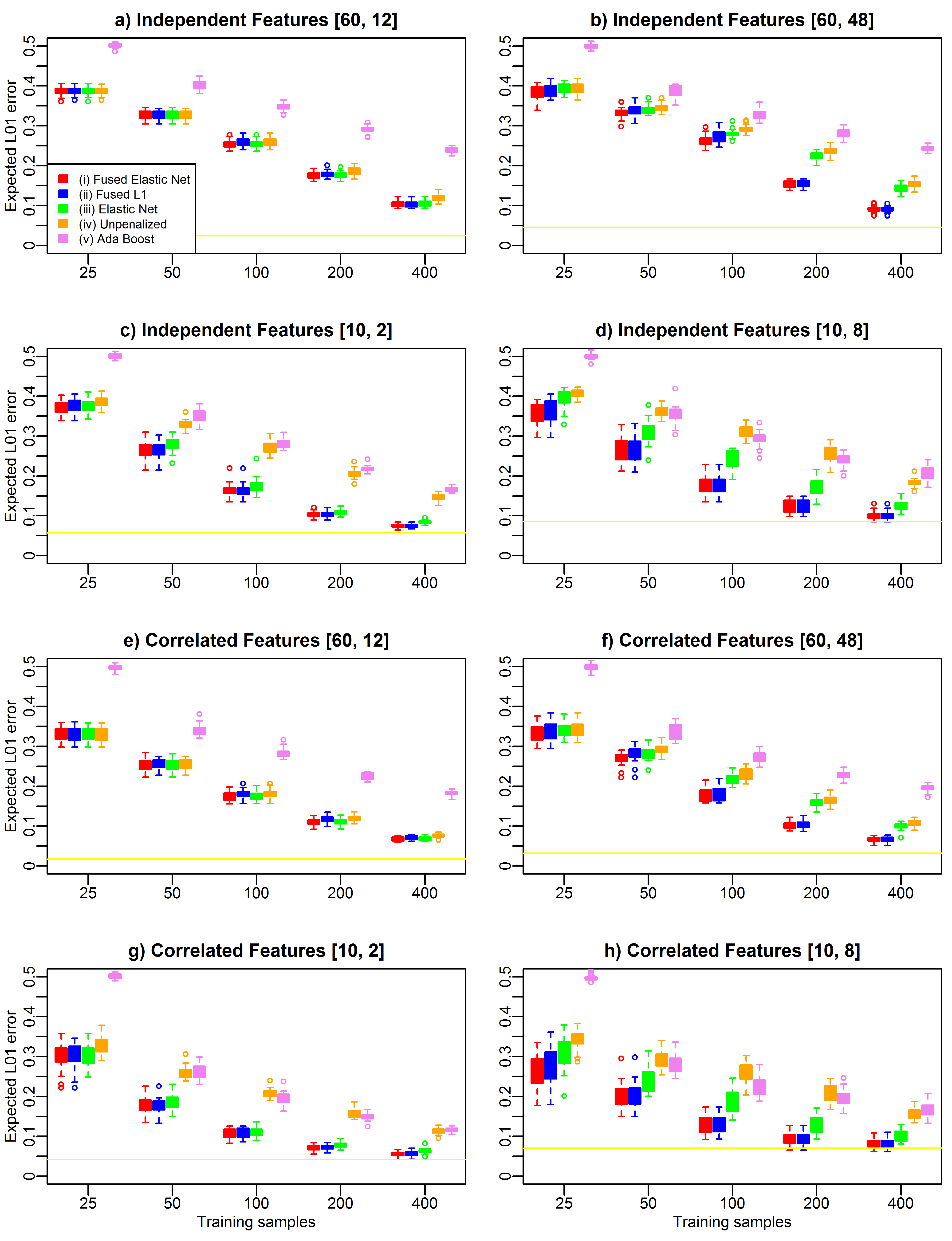}

\caption{Comparison between the expected $L_{01}$ errors of the five tested
models. The yellow horizontal lines represent the median of estimated
Bayes risks for the 20 quartet/data-set instances.\label{fig:Comparison-between-the}}
\end{figure}

The box-plots in \ref{fig:Comparison-between-the} depict the estimated
expected $L_{01}$ errors, each box representing the empirical distribution
of the error obtained from the corresponding 20 quartet/data-set instances
for a specific case (a-h) and specific amount of training examples.
The fused elastic net and the fused L1-penalized logistic regression
models dominate the three single-task models in the majority of cases
(cases b, c, d, f and h). This effect becomes significant in the case
of high similarity (b, d, f, h), particularly with slightly under-sampled
training data-sets (100 to 200 training samples). The fusing L1 penalty
seems to be less beneficial for the predictive performance in cases
of low task similarity with small training data-sets and/or non-sparse
true coefficient profiles (cases a, b with $\leq50$ training samples,
c with $\leq50$ training samples, e, f with $\leq50$ training samples
and g) as well as when the training data-set is big enough for the
single-task models to approach the Bayes risk (cases g and h with
$400$ training samples). The fused logistic regression model (i)
can still be used in these cases with a meta-parameter tuning procedure
such as cross validation which would automatically set the fusing
parameter $\nu$ to zero. 

The benchmarks show only a slight predictive advantage of the fused
elastic net model (i) compared to the fused lasso model (ii), particularly
in case b and f. A thorough look of the simulation results revealed
that for the majority of data-set instances the tuning of $\lambda_{2}$
has led to very low or zero values. 

\subsection{Comparison of Model Parameter Recovery\label{sub:Comparison-of-Model}}

\begin{figure}
\includegraphics[scale=0.75]{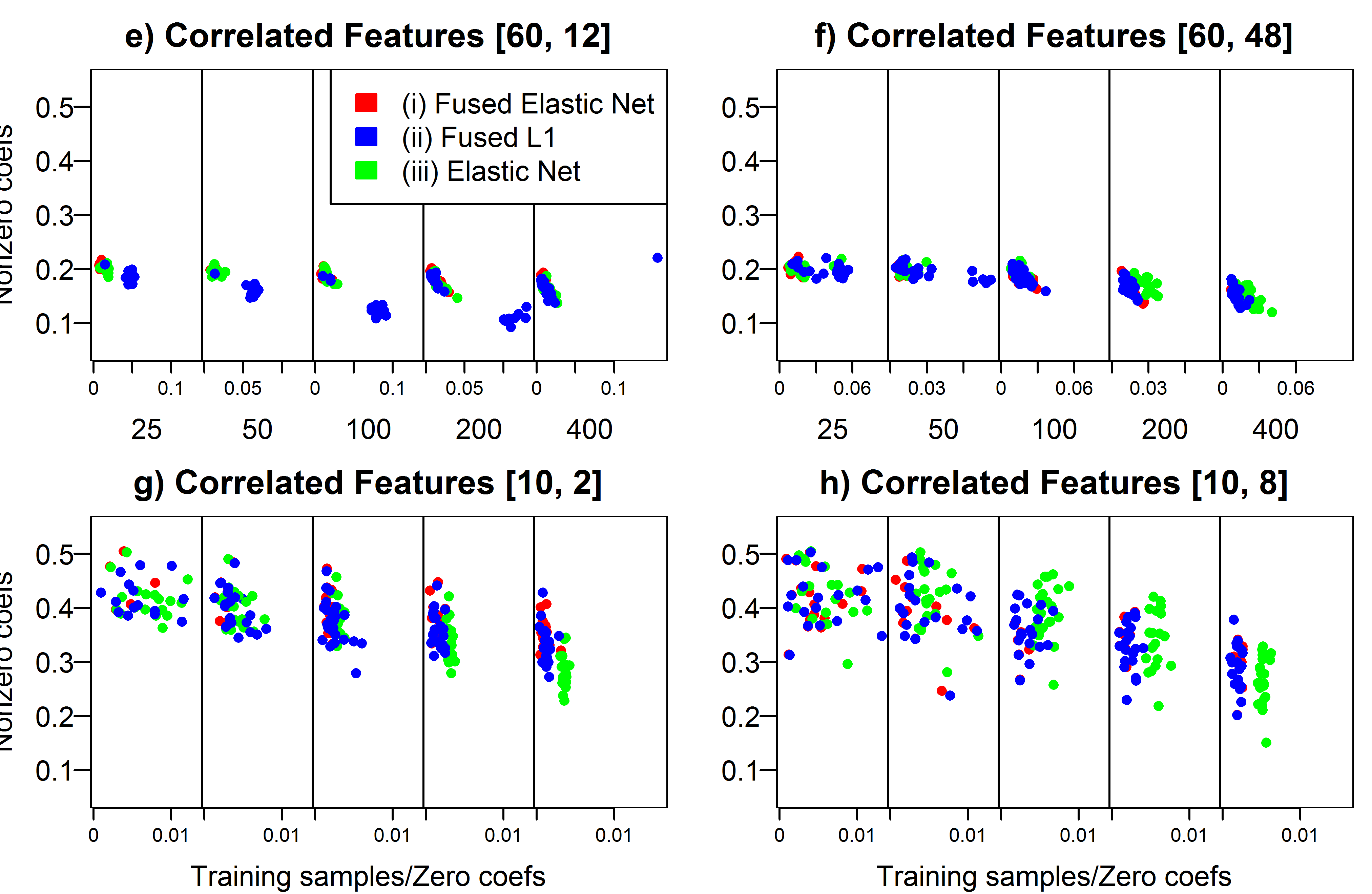}

\caption{Comparison between normalized Euclidean distances. The x-axis represents
the Euclidean distance between fitted model parameters and original
model parameters which are zero, divided by their count. The y-axis
represents the average Euclidean distance between fitted model parameters
and original model parameters which are non-zero, divided by their
count. \label{fig:Comparison-between-normalized}}
\end{figure}

In (\ref{fig:Comparison-between-normalized}), we plot the fitted
models for the ``Correlated features'' scenario according to their
normalized Euclidean distances with respect to the original model
parameters which are zero (x-axis) and non-zero (y-axis). In the sparse
cases, we notice that the distributions of the fused elastic net and
fused L1 models nearly overlap. The elastic net model favors some
higher discrepancy for the zero parameters, particularly in the cases
f with $\geq200$ training samples, g with $\geq200$ training samples
and h with $\geq100$ training samples. 

\section{DISCUSSION}

This work introduced the fused elastic net logistic regression for
multi-task binary classification. By means of sparsity inducing priors,
learning in this model enables to control sparsity for the individual
classifiers as well shared parameter patterns across related classification
tasks. Our results suggest that learning performance is enhanced if
a small but yet informative amount of data is available for the related
classification tasks. We observed no significant differences between
the performance between the fused L1 and elastic net logistic regression.
We hypothesize that this result will turn out in favor for the elastic
net variant if the sets of correlated variables also cover sets of
cardinality larger than two. The good overall classification performance
achieved in the considered simulation setting are encouraging for
real world applications, like in genome wide association studies in
biology where the data acquisition typically is costly and therefore
volume of data for individual classification tasks is notoriously
low. The ability to effectively leverage information across different
classification tasks will enable researchers to make progress in this
situation.

\bibliographystyle{plain}
\bibliography{mtlogreg_arxiv}

\newpage{}

\section*{APPENDIX}

\subsection*{Analytical expressions for the Hessian of $\tilde{f}^{\, k(j)}$}

\[
\nabla\nabla\tilde{f}^{\,(j)}(\chi_{\cdot j})=\left([\mathbf{w}]\odot X\right)^{T}\left([\mathbf{w}]\odot X\right)+(\boldsymbol{\lambda}_{2}+\boldsymbol{\rho})^{T}I,
\]
where we denoted

\[
\mathbf{w}:=\sqrt{\exp(-Y_{\cdot j}\odot X\chi_{\cdot j})}\div[\exp(-Y_{\cdot j}\odot X\chi_{\cdot j})].
\]

The symbol $'\div'$ denotes element-wise division between its vector
or matrix operands and $I$ denotes the identity matrix.

\subsection*{Analytical expressions for the gradient and hessian of the transformed
objective function $\phi^{k(j)}$ from Section 4.1}

\[
\tilde{X}:=X\div\left[\sqrt{\boldsymbol{\lambda}_{2}+\boldsymbol{\rho}}\right]^{T},
\]

\[
\tilde{\Omega}^{k}:=\Omega^{k}\div\left[\sqrt{\boldsymbol{\lambda}_{2}+\boldsymbol{\rho}}\right],
\]

\[
\boldsymbol{\Psi}^{k(j)}(\boldsymbol{\gamma}):=\exp\left(-Y_{\cdot j}\odot\tilde{X}\tilde{X}^{T}\boldsymbol{\gamma}-\rho Y_{\cdot j}\odot\tilde{X}\tilde{\Omega}_{\cdot j}^{k}\right),
\]

\[
\mathbf{w}_{j}(\boldsymbol{\gamma}):=\left(Y_{\cdot j}\odot\sqrt{\boldsymbol{\Psi}^{k(j)}(\boldsymbol{\gamma})}\right)\div\left(\mathbf{1+\boldsymbol{\Psi}}^{k(j)}(\boldsymbol{\gamma})\right),
\]

\begin{eqnarray*}
\nabla\phi^{k(j)}(\boldsymbol{\gamma}) & = & \tilde{X}\tilde{X}^{T}\left[\left(-Y_{\cdot j}\odot\boldsymbol{\Psi}^{k(j)}(\boldsymbol{\gamma})\right)\div\left(\mathbf{1+\boldsymbol{\Psi}}^{k(j)}(\boldsymbol{\gamma})\right)\right]+\tilde{X}\tilde{X}^{T}\boldsymbol{\gamma}
\end{eqnarray*}

\begin{eqnarray*}
\nabla\nabla\phi^{k(j)}(\boldsymbol{\gamma}): & = & \left(\mathbf{w}_{j}(\boldsymbol{\gamma})\odot\tilde{X}\tilde{X}^{T}\right)^{T}\left(\mathbf{w}_{j}(\boldsymbol{\gamma})\odot\tilde{X}\tilde{X}^{T}\right)+\tilde{X}\tilde{X}^{T}\boldsymbol{\gamma}
\end{eqnarray*}

\end{document}